# ENHANCED AUTOMATED QUALITY ASSESSMENT NETWORK FOR INTERACTIVE BUILDING SEGMENTATION IN HIGH-RESOLUTION REMOTE SENSING IMAGERY

*Zhili Zhang*[1], *Xiangyun Hu*[1,2 *], *and Jiabo Xu*[1]

[1]School of Remote Sensing and Information Engineering, Wuhan University, Wuhan 430079, P. R. China.
[2]Hubei Luojia Laboratory, Wuhan University, Wuhan 430079, P. R. China.

**ABSTRACT**

In this research, we introduce the enhanced automated quality assessment network (IBS-AQSNet), an innovative solution for assessing the quality of interactive building segmentation within high-resolution remote sensing imagery. This is a new challenge in segmentation quality assessment, and our proposed IBS-AQSNet allieviate this by identifying missed and mistaken segment areas. First of all, to acquire robust image features, our method combines a robust, pre-trained backbone with a lightweight counterpart for comprehensive feature extraction from imagery and segmentation results. These features are then fused through a simple combination of concatenation, convolution layers, and residual connections. Additionally, ISR-AQSNet incorporates a multi-scale differential quality assessment decoder, proficient in pinpointing areas where segmentation result is either missed or mistaken. Experiments on a newly-built *EVLab-BGZ* dataset, which includes over 39,198 buildings, demonstrate the superiority of the proposed method in automating segmentation quality assessment, thereby setting a new benchmark in the field.

***Index Terms***—Remote sensing imagery, interactive building segmentation, pre-trained backbone, deep learning, segmentation quality assessment.

## 1. INTRODUCTION

Annotation quality aSsessment (AQS) is a critical work in practical applications, focusing on identifying missed or mistaken areas in target or category annotations [1]. In this research, we conduct a segmentation quality assessment (SQA) study on the results from an interactive segmentation model, as illustrated in Figure 1. This study evaluates the interactive segmentation results of two types of prompts: point-based and bounding box-based. The ground truth of SQA highlights missed areas in green and mistaken areas in red. Given the demand for real-time and precise segmentation during interactive tasks, exploring an efficient, lightweight approach for SQA is both challenging and essential.

In the past decades, research on SQA has been limited, focusing primarily on two aspects: qualitative analysis of segmentation outcomes and the impact of sample quality on deep learning-based model. Qualitative evaluations typically employ empirical and rule-based designs. For example, methods like LabelMe [2] utilize annotation scoring functions based on control point counts, but these often overlook image context, leading to a certain unreliability. Vittayakorn and Hays [3] introduced more comprehensive scoring functions, incorporating factors such as annotation size and edge detection. The second aspect includes selecting high-quality samples for model training to improve their segmentation performance such as diverse labeling schemes [4], and neural network applications for map refinement [5]. Despite these efforts, a significant gap remains in the direct, quantitative evaluation of segmentation sample quality.

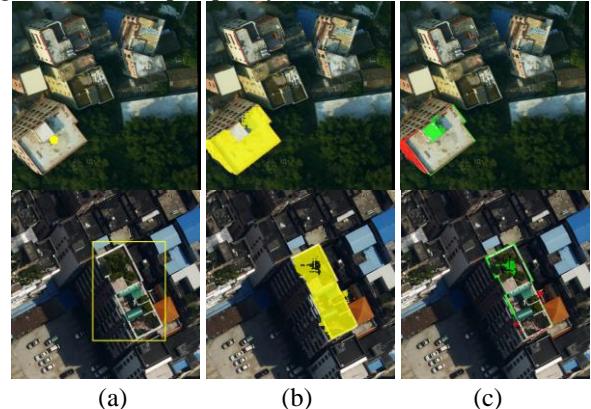

(a)        (b)        (c)

**Fig.1.** Case studies of quality assessment for interactive building segmentation: (a) images with point or bounding box prompts; (b) interactive segmentation results; (c) ground truth of SQA, with mistaken areas in red and missed areas in Green.

Our recent research focuses on assessing the quality of annotated samples for categories like buildings and water bodies in a single remote sensing image (RSI)[1]. This paper expands this scope to real-time SQA of interactive building segmentation, categorizing the results into missed and mistaken types at the instance level. Given the variable performance of interactive segmentation models such as Interformer [6], SAM [7], and PGR-Net [8] across different data sources and tasks, SQA for building segmentation encounters several challenges: conducting instance-level segmentation quality analysis, ensuring real-time predictive efficiency of the models, and maintaining robustness. To alleviate these, we employ the SAM model for building interaction segmentation. Our proposed approach utilizes SAM's backbone for robust feature extraction, complemented by a lightweight backbone for rapid feature extraction from images and segmentation results. These features are fused through techniques like convolutional layers, concatenation, sampling, and residual concatenation. The SQA result is then predicted by using a multiscale feature difference quality assessment module. The proposed IBS-AQSNet is able to realize a fast and stable quality assessment of the interaction segmentation results.

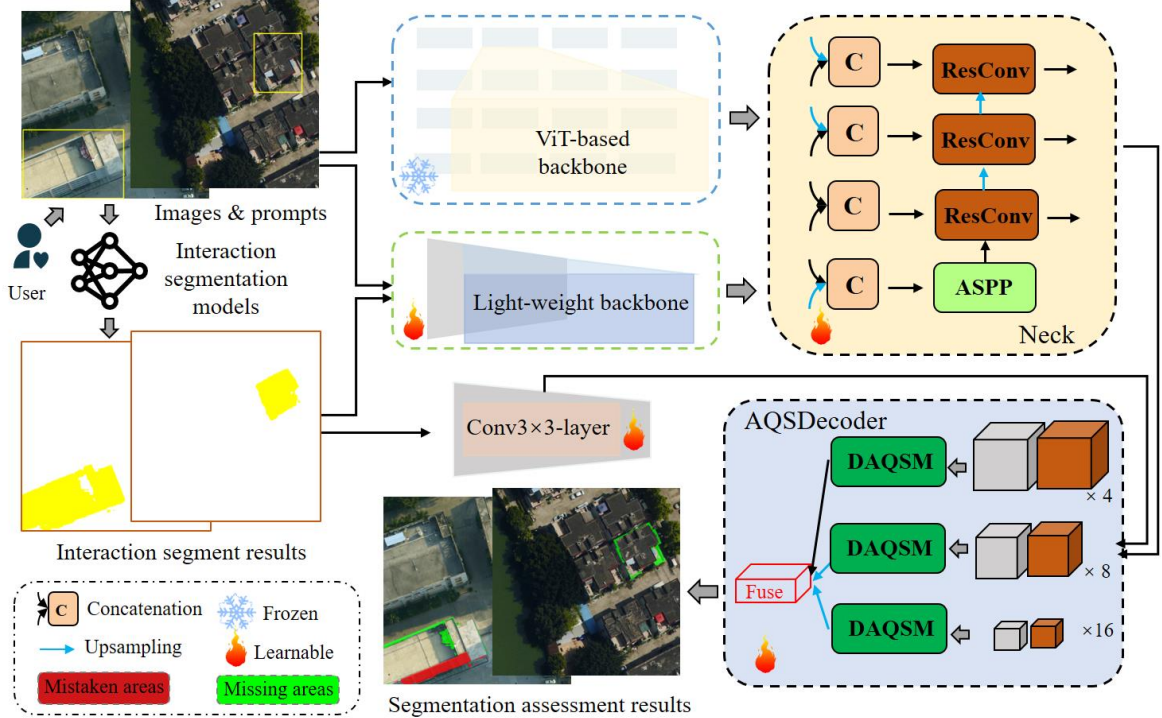

**Fig.2.** The flowchart of the proposed IBS-AQSNet.

## 2. METHODOLOGY

In this section, the proposed IBS-AQSNet is described in detail. The flowchart of the proposed IBS-AQSNet is shown in Figure 2. Firstly, the segmentation results and their image features provided by the interactive segmentation model are used as inputs to the IBS-AQSNet. The proposed method first extracts the features of the segmentation result by taking the segmentation result and image as the input of a lightweight backbone. Then image and segmentation features are simply fused by convolution, sampling and concatenation operations. Finally, a multiscale differential quality assessment decoder is utilized to obtain the quality assessment results.

### 2.1. Robust feature extraction and feature fusion

Figure 2 illustrates our method employs dual backbones for robust image feature extraction: a Vision Transformer (ViT)-based backbone from SAM-b, trained on extensive datasets for robust feature extraction [7], and a modified, lightweight backbone based on ResNet18 [9]. The ViT-based backbone is utilized for extracting rich semantic features from images across four stages, without involvement in training. In contrast, the ResNet18-based backbone is adapted for rapid learning on interaction segmentation results and images. This adaptation includes expanding the input channels from 3 to 4 and removing the classifier, thus retaining spatially detailed feature maps across the four stages. This modified backbone facilitates easier training and ensures stable feature map integration with the pre-trained ViT-based backbone.

The process involves fusing feature maps from both backbones across their respective four stages. To align spatial resolutions, the first three stages of the ViT-based backbone's feature maps are resized to correspond with those from ResNet18. Meanwhile, the fourth-stage feature maps of ResNet18 are modified to align with the fourth stage of the ViT-based backbone. To keep pre-trained features from the ViT-basd backbone, a 1×1 convolution kernel is employed to standardize the channel dimensions to those of ResNet18 across all four stages. Following this, the feature maps from each stage are concatenated in a sequential manner. Specifically, the fourth-stage fused feature maps are processed using ASPP [10], then subjected to stage-by-stage up-sampling and concatenation with the preceding stage's feature maps, along with processing via convolutional layers incorporating residual concatenation. This simple strategy results in the generation of fused feature maps at three distinct scales, as depicted in the 'Neck' of Figure 2. Aligning with our recent work [1], the last layer of the fused features is directed to the classifier as an auxiliary output. The final fused feature maps maintain sizes of 1/4, 1/8, 1/16, and 1/16 relative to the original image, with respective channel counts of 64, 128, 256, and 512.

### 2.2. The AQS Decoder

Following our recent AQSNet [1], we continuously use this advanced multi-scale feature difference quality assessment module, referred to as the AQS decoder, for assessing segmentation results. The process begins with rapid feature learning using two convolutional layers, with a 3×3 kernel and a stride of 2. The next step involves subtracting the fusion features from the segmentation features across three scales. At each stage, the CSAM module from AQSNet is utilized for an in-depth analysis of these differentiated features, thereby enhancing the detection of missed and mistaken areas in the segmentation results. Subsequently, these enhanced differential features are uniformly resized to align with the scale of the largest feature map. The culmination of this process involves integrating these concatenated features into a classifier, which is adept at pinpointing both missed and incorrect areas within the segmentation results.

### 2.3. Loss fuction and accuracy assessment

During the training phase, the supervision of the auxiliary output and SAQ results is accomplished using two loss functions: the cross entropy loss function (CE) and the dice loss function [11]. The loss computation formula for the segmentation result (Ouput) is as follows:

$$loss = \gamma_1 * loss_{CE}(Ouput, GT) + \gamma_2 * loss_D(Ouput, GT) \quad (1)$$

Here, the parameters $\gamma_1$ and $\gamma_2$ are set to 0.5, the *GT* means the corresponding ground truth, and $loss_D$ means the dice loss function.

Furthermore, we employ four common metrics to evaluate the accuracy of our proposed method: F1-score, recall, precision, and overall accuracy (OA). These metrics are determined by comparing the prediction map against its ground truth (GT), and involve the calculation of true positives (TP), false positives (FP), true negatives (TN), and false negatives (FN). The corresponding equations for these metrics are presented in equations (2) - (5).

$$Precision = \frac{TP}{TP + FP} \quad (2)$$

$$Recall = \frac{TP}{TP + FN} \quad (3)$$

$$F1 = \frac{2 * Precision * Recall}{Precision + Recall} \quad (4)$$

$$OA = \frac{TP + TN}{TP + FP + TN + FN} \quad (5)$$

**Table 1.** Quantitative analysis of IBS-AQSNet under different experimental setups on the EVLab-BGZ. PIF (using pre-trained image features from SAM-b's backbone), AQSD (the AQS decoder), M (Million), G (Billion).

| Methods | Params (M) | Flops (G) | Missed-areas | | | Mistaken-areas | | | OA |
|---|---|---|---|---|---|---|---|---|---|
| | | | Precision | Recall | F1-score | Precision | Recall | F1-score | |
| Baseline | 32.66 | 81.05 | 34.940 | **70.326** | 46.685 | **64.691** | 52.308 | 57.844 | 98.370 |
| Baseline + PIF | 43.12 | 99.06 | 51.196 | 62.748 | 56.386 | 55.497 | 60.698 | 57.981 | 98.934 |
| Baseline + PIF + AQSD | 41.62 | 95.70 | **51.376** | 63.734 | **56.892** | 59.383 | **61.132** | **60.245** | **98.951** |

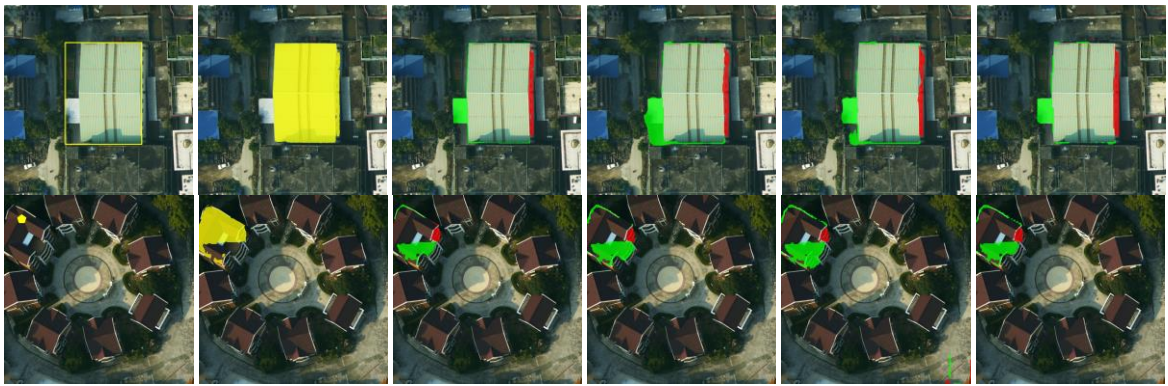

**Fig.3.** IBS-AQSNet performance visualization on EVLab-BGZ under different experimental setups. (a) images & prompts, (b) interactive segmentation results, (c) SQA ground truth, (d) Baseline, (e) Baseline with PIF, (f) Baseline with PIF and AQSD. Key: Red for mistaken areas, Green for missed areas.

## 3. EXPERIMENTAL RESULTS AND ANALYSIS

To evaluate the effectiveness of our proposed IBS-AQSNet, we create a novel instance-level building extraction dataset, called *EVLab-BGZ*, primarily located in Guangzhou city. This dataset includes both aerial and satellite images with resolutions of 0.2 and 0.5, captured in 2019, each measuring $512 \times 512$ pixels (refer to Figure 4). It comprises 2,825 aerial and 1,006 satellite images, encompassing a total of 39,198 building instances. In these buildings, each building is labeled independently and multiple buildings are rarely labeled together. The dataset is partitioned into 3,000 images for training and 831 for testing. Additionally, we have made this dataset publicly available. The building density within the dataset varies significantly, ranging from a single building to as many as 51 buildings per image, with each building occupying over 2,500 pixels. Given these attributes, the dataset presents a considerable challenge for building instance segmentation tasks.

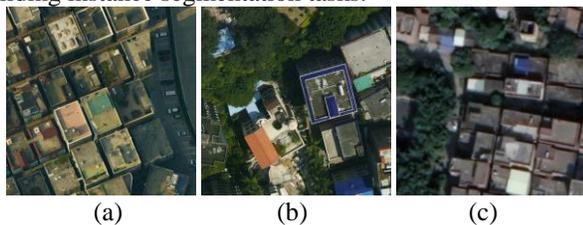

(a)       (b)       (c)

**Fig.4.** Some building samples from EVLab-BGZ: (a-b) Aearial imagery; (b) Satelite imagery.

The designed methods are validated on the EVLab-BGZ. Both quantitative and qualitative analyses are conducted to evaluate their performance. As shown in Table 1, the baseline, which incorporates a ResNet18 backbone and an UperNet decoder [12], achieves an F1 score of 46.685% for missed areas and 57.844% for mistaken areas, with an OA of 98.370%. These findings highlight the relative simplicity in recognizing background areas in SQA, in contrast to the more complex task of identifying missed and mistaken areas. The use of pre-trained image features leads to a significant 9.71% increase in the F1 score for missed areas, while only marginally improving the detection performance for mistaken areas. The precision for mistaken areas declines, and although recall increases, the F1 score remains unchanged, suggesting possible underutilization of the used pre-training features. The introduction of AQS decoder gives an improvement in detecting both mistaken and missed areas, indicating the necessity for specially designed modules in tasks such as building SQA.

Figure 3 presents the visualization of our proposed methods through two prompts for building interactive segmentation, illustrating that our method closely aligns with the SQA ground truth. Columns (d) - (f) of Figure 3 depict that our method, utilizing the AQS decoder, more accurately identifies missed regions than the baseline model, which often misidentifies such areas. The precision of the baseline is further enhanced by the addition of pre-training features. In point prompt interaction segmentation, our approach demonstrates similar enhancements. These findings underscore the effectiveness of our method in enabling automatic quality assessment for building interaction segmentation. We have integrated this approach into our interaction segmentation software, enhancing its capacity for innovative building interactions and delivering superior segmentation outcomes. More visualization results are available here[1].

## 4. CONCLUSION

This paper introduces the IBS-AQSNet, a noval network designed for segmentation quality assessment (SQA) of interactive building segmentation in high-resolution remote sensing imagery. The proposed IBS-AQSNet integrates a robust, pre-trained backbone with a lightweight counterpart, enabling efficient and comprehensive feature extraction. The network employs a straightforward fusion of these features and integrates a multi-scale differential quality assessment decoder, adept at identifying errors, including missed and mistaken areas, in SQA results. The proposed method effectively addresses the challenge of building segmentation quality assessment. Experimental validation on the newly built EVLab-BGZ dataset confirms that the IBS-AQSNet excels in the building SQA task, representing a progression in remote sensing image analysis.

## 5. ACKNOWLEDGMENT

This work was supported by the Special Fund of Hubei Luojia Laboratory under grant 220100028 and 230700006, and the Fundamental Research Funds for the Central Universities, China under grant 2042022dx0001.

---

[1] https://github.com/zhilyzhang/IBS-AQSNet